\documentclass[lettersize,journal]{IEEEtran}
\pdfoutput=1
\usepackage{amsmath,amsfonts}
\usepackage{amsfonts,amssymb}
\usepackage{algorithmic}
\usepackage{algorithm}
\usepackage{array}
\usepackage[caption=false,font=normalsize,labelfont=sf,textfont=sf]{subfig}
\usepackage{textcomp}
\usepackage{stfloats}
\usepackage{url}
\usepackage{verbatim}
\usepackage{graphicx}
\usepackage{cite}
\usepackage{enumitem}
\usepackage[numbers,sort&compress]{natbib}
\usepackage{bm}
\usepackage{color}
\usepackage{amssymb}
\usepackage[justification=centering]{caption}
\usepackage{setspace}
\makeatletter
\renewcommand{\maketag@@@}[1]{\hbox{\m@th\normalsize\normalfont#1}}
\newcommand{\tabincell}[2]{\begin{tabular}{@{}#1@{}}#2\end{tabular}}
\makeatother

\begin{document}
	\title{\large Privacy-Preserving Joint Edge Association and Power Optimization for the Internet of Vehicles via Federated Multi-Agent Reinforcement Learning }
	
	\author{Yan~Lin,~\IEEEmembership{Member,~IEEE},
		Jinming~Bao,  
		Yijin~Zhang,~\IEEEmembership{Senior Member,~IEEE}, Jun~Li,~\IEEEmembership{Senior Member,~IEEE}, Feng~Shu,~\IEEEmembership{Member,~IEEE} and Lajos~Hanzo,~\IEEEmembership{Life Fellow,~IEEE}
		}
	\maketitle

	\begin{abstract}
Proactive edge association is capable of improving wireless connectivity at the cost of increased handover (HO) frequency and energy consumption, while relying on a large amount of private information sharing required for decision making. In order to improve the connectivity-cost trade-off without privacy leakage, we investigate the privacy-preserving joint edge association and power allocation (JEAPA) problem in the face of the environmental uncertainty and the infeasibility of individual learning. Upon modelling the problem by a decentralized partially observable Markov Decision Process (Dec-POMDP), it is solved by federated multi-agent reinforcement learning (FMARL) through only sharing encrypted training data for federatively learning the policy sought. Our simulation results show that the proposed solution strikes a compelling trade-off, while preserving a higher privacy level than the state-of-the-art solutions.
	\end{abstract}
\vspace{-0.3cm}
	\section{Introduction}
	As a promising relative of the Internet-of-Things (IoT), the Internet of Vehicles (IoV) is capable of supporting delay-sensitive services for improving the road safety, traffic efficiency, autonomous driving and real-time information interaction in intelligent transportation systems (ITSs) \cite{1}. In the IoV, each vehicle is typically connected to the infrastructure, to other vehicles, pedestrians or networks under the vehicle-to-everything (V2X) paradigm. Pioneered by the Google car concept, vehicles have communications, storage and learning capabilities and make their own decisions for supporting ultra-high reliability and low latency communication (URLLC) services \cite{2} \cite{9525439}. 
	
	To satisfy the resultant connectivity requirement, edge association through access points (APs), such as road side units (RSUs), becomes particularly essential under the ever-increasing traffic encountered \cite{4} \cite{7378276}. Inevitably, the inherent mobility of the IoV results in frequent handovers (HOs), and hence in throughput reduction, call dropping as well as additional energy dissipation \cite{5}. Moreover, in order to response the call for energy conservation and carbon reduction, the transmit power of RSUs has to be accurately controlled to meet both the data rate and energy consumption requirements of V2X communication \cite{9618822}. Therefore, edge association and power allocation have to be jointly considered in the IoV to support URLLC services.
	
 In view of the fact that the joint edge association and power allocation (JEAPA) problem of the IoV is typically treated as a sequential decision-making problem in the face of vehicular mobility and channel states uncertainty, reinforcement learning (RL) can be employed for formulating good policies by learning from the interactions with the environment. For instance, Khan \textit{et al.} of \cite{8} adopted a distributed RL framework for edge association, which meets the transmission rate requirements while minimizing the network coordination overhead. In our previous work~\cite{9}, we developed a deep RL (DRL) based edge association scheme for striking a trade-off between the connectivity and HO rate of the heterogeneous IoV. However, both of them rely on a large amount of information exchange and sharing in a centralized way, which potentially increases the risk of privacy leakage, concerning their location and social data.
	
	In order to reduce the information sharing required by centralized processing, multi-agent RL (MARL) is developed for our decision-making system, where the agents learn to make their own decisions cooperatively through their local observations for the same global reward \cite{10}. As a further advance, to facilitate the decentralized training of agents, Konecny \textit{et al.} \cite{11} proposed federated learning (FL) for guaranteeing training data on edge devices rather than centrally. Motivated by the benefits that local training data is not uploaded and shared, a number of researchers have exploited FL in privacy preservation in the context of DRL-based decision-making problems \cite{12, 9205252, 13}. The existing literature typically adopts DRL to train the policy used for the resource allocation, and averages the weights of the agents' Deep Neural Networks (DNNs) at the APs to generate a joint policy for the next iteration of the local training. Although the individual state-information of each agent can be stored locally with Gaussian encryption, the aggregated DNN weights have to be shared amongst the APs, which may cause privacy leakage, as demonstrated by the model inversion attacks of \cite{fredrikson2015model}. Moreover, the structure of DNNs used for different agents may be different, which makes the process of weights aggregation hard to implement in practice. As a further advance, a novel FL assisted MARL system is investigated in \cite{14}, where the agents train their policies centrally by only sharing the encrypted outputs of the DNNs, instead of the aggregated DNN weights. More explicitly, the outputs of DNNs, that can approximate the state-action-value (Q-value) function, contain substantial private information, which is more beneficial for the model training than for the shared aggregated DNN weights.
	
    Against the above backdrop, we conceive a federated MARL (FMARL) based JEAPA solution, where all vehicular agents federatively learn their policies through only sharing the encrypted local Q-values for centralized training and make decisions distributively relying on their own local observations. \textit{To the best of our knowledge, this is the first attempt in the open literature to study the privacy-preserving JEAPA problem of the IoV relying on a FMARL framework.} Our main contributions are boldly and explicitly contrasted to the literature in Table \ref{table} and are detailed as follows:
	\begin{itemize}
		\item{We conceive a federated multi-agent JEAPA framework for vehicular mobility and channel states uncertainty, with the aim of improving the long-term trade-off involving the connectivity, the HO overhead and the energy consumption while preserving the privacy.}
		\item{We propose a privacy-preserving-based JEAPA solution under our federated multi-agent framework, which shares the encrypted local Q-values for federatively learning their policies. In particular, even though some vehicular agents cannot learn individually, they are capable of making decisions distributively with the aid of federative training results.}
		\item{Our numerical simulation results show that the proposed solution outperforms the state-of-the-art benchmarks, in terms of its convergence, HO-rate reduction, and connectivity improvement with the additional benefit of privacy preservation. Moreover, the trade-off between the convergence and the privacy protection levels is also quantified.}
	\end{itemize}
\begin{table*}[!t]
	\centering
	\vspace{-0.5cm}
	\caption{Related Contributions\\
		( 
		$\textbf{\footnotesize A}:\text{\footnotesize Global state}$,
		$\textbf{\footnotesize B}:\text{\footnotesize Local state}$,
		$\textbf{\footnotesize C}:\text{\footnotesize Aggregated DNN weights}$,
		$\textbf{\footnotesize D}:\text{\footnotesize Encrypted local Q-values.}$)}
	\label{table}
	\renewcommand\arraystretch{1.1} 
	\footnotesize
	\begin{tabular}{|l|c|c|c|c|c|c|c|c|c|c|}
		\hline
		& \tabincell{c}{\cite{4} \\ $\text{-2021}$}
		& \tabincell{c}{\cite{5} \\ $\text{-2019}$}
		& \tabincell{c}{\cite{9618822} \\ $\text{-2022}$}
		& \tabincell{c}{\cite{8} \\ $\text{-2019}$}
		& \tabincell{c}{\cite{9} \\ $\text{-2020}$} 
		& \tabincell{c}{\cite{10} \\ $\text{-2019}$}
		& \tabincell{c}{\cite{12} \\ $\text{-2020}$}
		& \tabincell{c}{\cite{9205252} \\ $\text{-2021}$} 
		& \tabincell{c}{\cite{13} \\ $\text{-2021}$}
		& $\textbf{Proposed}$
		\\  
		\hline
		Vehicular network & & & $\checkmark$ & $\checkmark$ & $\checkmark$ & $\checkmark$ &  & & & $\checkmark$\\  
		\hline
		Unknown user mobility & & $\checkmark$ & & $\checkmark$ & $\checkmark$ &$\checkmark$  & &  & $\checkmark$ & $\checkmark$\\ 
		\hline
		JEAPA & & & & & & &  & & & $\checkmark$\\ 
		\hline
		Data rate & $\checkmark$ &  & $\checkmark$& $\checkmark$ & $\checkmark$ & $\checkmark$ & $\checkmark$ & $\checkmark$ & &  $\checkmark$\\ 
		\hline
		HO overhead  & & $\checkmark$ & & & $\checkmark$ & & & & &  $\checkmark$\\ 
		\hline
		Energy consumption  & $\checkmark$ &  &$\checkmark$ & &  & & &$\checkmark$ &$\checkmark$ &$\checkmark$ \\ 
		\hline
		Multi-agent system  &  & & & &  &$\checkmark$ & & &$\checkmark$  &$\checkmark$ \\ 
		\hline
		Information sharing & $\textbf{A}$ & $\textbf{A}$ &$\textbf{A}$ &$\textbf{B}$ &$\textbf{A}$& $\textbf{B}$& $\textbf{C}$& $\textbf{C}$& $\textbf{C}$+$\textbf{D}$ & $\textbf{D}$\\
		\hline
	\end{tabular}
	\vspace{-0.5cm}
\end{table*}
\vspace{-0.1cm}
	\section{System Model And Problem Formulation}
	In this section, the system model and the problem formulation are introduced, respectively. 
\vspace{-0.25cm}
	\subsection{System Model}
	We consider a typical IoV network consisting of $K$ vehicles and $R$ RSUs. The vehicles drive along a twin-lane freeway, indexed by ${\cal K} \triangleq \{ 1,2, \ldots ,K\}$, which communicate with the RSUs using orthogonal resource blocks to mitigate the inter-user interference. The RSUs, indexed by ${\cal R} \triangleq \{ 1,2, \ldots ,R\}$, are evenly distributed on both sides of the freeway to provide high-rate services. The macro base station (MBS) is deployed for providing always-on coverage and serving as a central data-processing point. The system has a time-slot (TS) index set of ${\cal T} \triangleq \{ 1,2, \ldots ,T\}$, where both the channel state information (CSI) and the system parameters remain unchanged during each TS, but may vary randomly across different TSs.
	
	We assume that vehicle $k\in {\cal K}$ can only communicate with the RSUs within a limited coverage range and select one of the RSUs to be associated with at TS $t\in {\cal T}$. Let us denote the maximum number of observable RSUs as $O_{\max}$ and define the edge association indicator vector between vehicle $k$ and all RSUs as $\boldsymbol{c}_t^k = [c_t^{k,1}, \ldots ,c_t^{k,R}]$.
    Explicitly, $c_t^{k,r}=1$ if RSU $r\in{\cal R}$ is associated with vehicle $k$ at TS $t$, and $c_t^{k,r} = 0$ otherwise. If the association changes during a pair of adjacent TSs, an HO is triggered for vehicle $k$ at TS $t$, given by ${H_t^k} =\boldsymbol{1}_{\{\boldsymbol {c}_t^k \ne \boldsymbol{c}_{t-1}^k\}}$, where $\boldsymbol{1}_{\{\cdot\}}$ equals to $1$, if the condition is satisfied and $0$ otherwise.
	
	The transmit power of RSUs can be selected from $P$ levels in $[P_{\min}, P_{\max}]$. As such, the power allocation indicator vector of vehicle $k$ at TS $t$ is given by $\boldsymbol{e}_t^k= [e_t^{k,1}, \ldots ,e_t^{k,P}]$, where if the ${k^{th}}$ vehicle selects the power level $p$ for its associated RSU at TS $t$, we have $e_t^{k,p} = 1$, and $e_t^{k,p} = 0$ otherwise. Let $P_t^{k,r}$ denote the transmit power of RSU $r$ associated with vehicle $k$ at TS $t$, yielding $e_t^{k,p}=\boldsymbol{1}_{\{c_t^{k,r}=1, p=P_t^{k,r}\}}$.
	
	In our assumption, all transceivers are equipped with a single antenna, and we only take the small-scale fading and the path loss into consideration. Given that vehicle $k$ is associated with RSU $r$ at TS $t$, the achievable downlink data rate of vehicle $k$ can be represented as:
		\begin{equation}
			{Rate_t^k}= {\log _2}(1 + \frac{{\mathop  P_t^{k,r}G_t^{k,r}}}{{\sigma _0^2}}),
		\end{equation}
	where $G_t^{k,r}$ is the channel gain between vehicle $k$ and RSU $r$ at TS $t$. We assume that the additive Gaussian white noise (AWGN) has zero mean and identical variance $\sigma _0^2$ at all the vehicles. Additionally, the minimum data rate $R_{\min}$ required by all vehicles at each TS is assumed to be the same.
	
\vspace{-0.1cm}
	\subsection{Problem Formulation}
	The aim of our optimization problem is to maximize the long-term per-user trade-off between the connectivity versus the cost quantified in terms of the number of HOs and the associated RSU's transmit power consumption. Similar to~\cite{10}, we define a normalized trade-off utility function for our JEAPA problem at TS $t$, which can be formulated as
		\begin{equation}
			\begin{array}{l}
				{U_t^k}={\omega _1}\frac{{Rate_t^k}}{R_{\min}}  -  {\omega _2}{H_t^k} -  
				{\omega _3}\frac{{P_t^{k,r}}}{{{P_{\max}}}}.
			\end{array}
		\end{equation} 
	Herein, ${\omega _1},{\omega _2},{\omega _3} \in [0,1]$ quantify the weighting factor assigned to the connectivity benefit, HO overhead and transmit power of RSU, respectively. 
	
	Subject to the minimum transmit rate constraint, our problem can be formulated as
		\begin{subequations}\label{minPC}
			\begin{align}
				\max_{\boldsymbol{c}_t^k,\boldsymbol{e}_t^k}~
				& {\mathbb{E}}[\mathop \frac{1}{K}\sum\limits_{t = 1}^T {\sum\limits_{k = 1}^K{{U_t^k}}] }\\\label{a}
				s.t.~
				&{\sum\limits_{k =1}^K {c_t^{k,r}} \le 1 ,\forall r \in {\cal R},\forall t \in {\cal T}},\\ \label{b}
				&{\sum\limits_{r =1}^R {c_t^{k,r}} = 1 ,\forall k \in {\cal K},\forall t \in {\cal T}},\\\label{c}
				& {{Rate_t^k} \ge {R_{\min }},\forall k \in {\cal K},\forall t \in {\cal T}}.
			\end{align}
		\end{subequations}
	Herein, (\ref{a}) indicates that each RSU can only serve at most one vehicular user simultaneously and (\ref{b}) guarantees seamless connectivity, while (\ref{c}) reflects the minimum data rate requirement.
	
	It can be observed that problem (\ref{minPC}) is a sequential dynamic decision-making problem to be optimized over multiple TSs. In view of the stochastic environmental states represented by the vehicular mobility, conventional optimization techniques, such as convex optimization and linear programming, cannot be readily applied. Although DRL is widely exploited for constructing policies to achieve certain long-term average objectives \cite{15}, relying on large amounts of private information interaction and sharing in a single-agent framework is still impractical. To this end, we adopt our FMARL technique for solving the privacy-preserving JEAPA problem in a decentralized framework.
	\begin{figure*}[!t]
		\centering
		\includegraphics[width=0.8\textwidth]{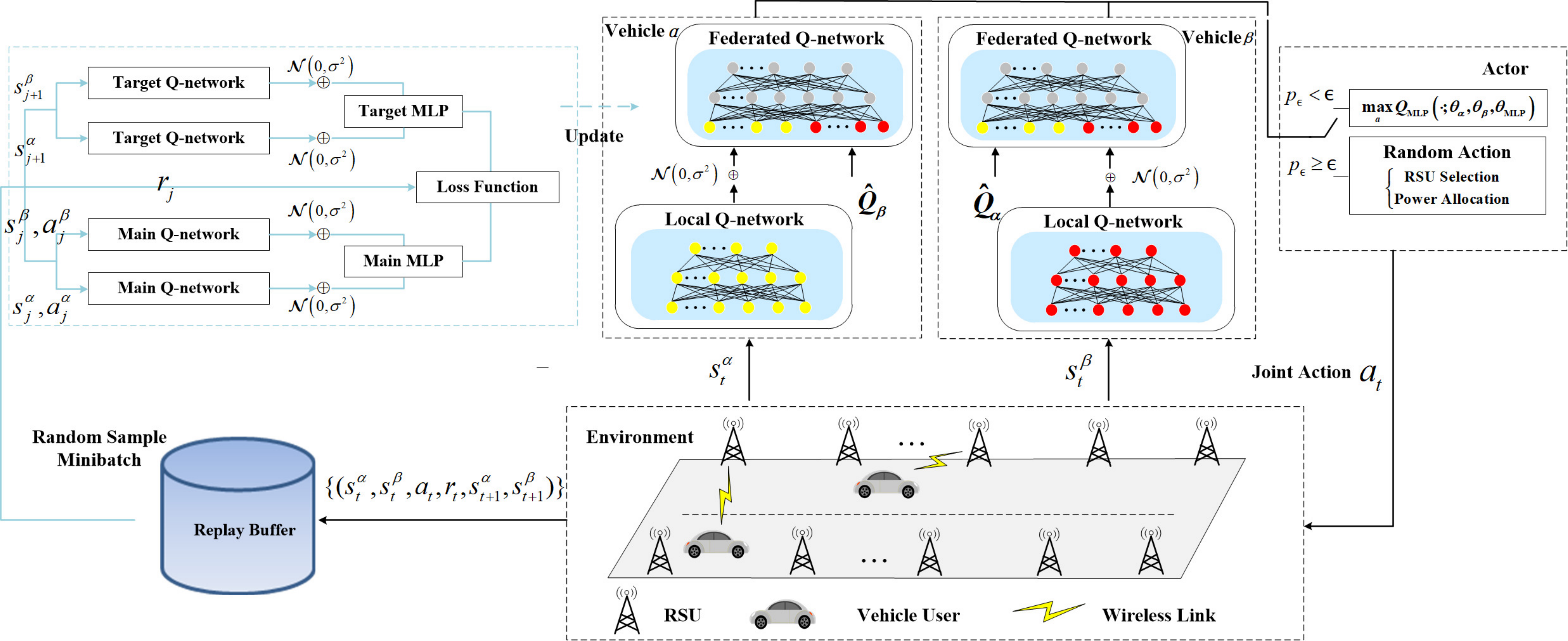}
		\caption{The framework of federated multi-agent JEAPA.}
		\label{fig_1}
\vspace{-0.5cm}
	\end{figure*}

	\section{Privacy-Preserving Multi-Agent Joint Edge Association And Power Allocation Solution}
	In this section, we first model the JEAPA problem as a decentralized partially observable Markov Decision Process (Dec-POMDP). Operating in the face of uncertainty, we resort to the FMARL framework for developing a novel privacy-preserving JEAPA solution. 
\vspace{-0.3cm}
	\subsection{Dec-POMDP Design}
    Intuitively, given the fact that the state-information cannot be fully observed by vehicular agents and both the vehicular mobility and channel states are unknown in advance, the JEAPA problem can be constructed as a Dec-POMDP problem in that all vehicles act as agents to make decisions individually relying on their own local observations. The Dec-POMDP problem can be modeled as
	\begin{enumerate}[fullwidth,itemindent=1em,label=\arabic*)]
		\item{\emph{Observations}: For the vehicular agent $k\in\mathcal{K}$, its observation at TS $t$ may be defined as	
					$\boldsymbol{o}_t^k = [
					\boldsymbol{G}_t^k, \boldsymbol{L}_t^k,
					L_{t-1}^{k,r}]$,
			where
			\begin{itemize}
				\item{
					$\boldsymbol{G}_t^k \! = \! [G_t^{k,1}, \ldots, G_t^{k,O_{\max}}]$ is the set of CSIs between vehicle $k$ and its observable RSUs at TS $t$;}
				\item{$\boldsymbol{L}_t^k \! = \! [L_t^{k,1}, \ldots, L_t^{k,O_{\max}}]$ is the set of the RSUs' locations observed by vehicle $k$ at TS $t$;}
				\item{$L_{t-1}^{k,r}$ is the location of RSU $r$ associated with vehicle $k$ at TS $t-1$.}
			\end{itemize}
		}
		\item{\emph{Actions}: According to the decision policy, each agent has to select the associated RSU and configure its transmit power level, simultaneously. Thus, for the vehicular agent $k\in\mathcal{K}$, its action at TS $t$ may be defined as
$\boldsymbol{a}_t^k =  [
					\boldsymbol{c}_t^k,
					\boldsymbol{e}_t^k].
$
		}
		\item{\emph{Reward}: Provided that vehicular agent $k$ takes action $\boldsymbol{a}_t^k = \boldsymbol{a}$ when $\boldsymbol{o}_t^k= \boldsymbol{o}$ at TS $t$, the system will receive a global reward $r_t$. Since the objective of problem (\ref{minPC}) is to maximize the long-term system utility function, we design the per-user average trade-off (PAT) as the global reward. Moreover, when the constraints (\ref{a})-(\ref{c}) are not satisfied, a penalty term $\rho_t$ is added on the PAT. Then, we have
$r_t = \frac{1}{K}\sum\limits_{k=1}^K {{U_t^k}} + \rho_t$.
	}
	\end{enumerate}

In practical multi-agent IoV scenarios, vehicles can observe their own real-time locations and speed based on their pre-installed sensors and positioning technology. However, they may not have timely or accurately reward feedback due to authority or trust issues. To deal with this impediment, we classify the vehicular agents into a pair of types, namely $\alpha \in \cal K$  and $\beta \in \cal K \backslash \{ \alpha \}$ according to the availability of reward knowledge: 
\begin{itemize}
	\item{\emph{Type-$\alpha$ vehicular agents}: They can observe their local states, and obtain the corresponding global reward in a timely and accurate manner;}
	\item{\emph{Type-$\beta$ vehicular agents}: They can observe their local states, but the global reward cannot be obtained due to reasons of privacy preservation.}
\end{itemize}

	\subsection{Problem Reformulation}
	Based on the Dec-POMDP constructed, we define the JEAPA policy $\pi$ as the mapping from the current observations to a series of actions. To maximize the expected long-term global reward, the Q-value function is adopted to evaluate a single action at a state, defined as   
	\begin{equation}
		Q^{\pi}(\boldsymbol{o}, \boldsymbol{a}) = \mathbb{E}[\sum_{l=0}^{T-t}\gamma^{l}r_{t+l}|\boldsymbol{o}_t= \boldsymbol{o},\boldsymbol{a}_t= \boldsymbol{a}],
	\end{equation}
	where $\gamma \in [0,1]$ is a discount factor that reflects the effect of future rewards on the optimal policy. 
	
	To satisfy the privacy-preserving requirements, we adopt the Gaussian differential method of \cite{16} to encrypt the shared local Q-values amongst vehicular agents, which can be defined as 
	\begin{equation} \label{5}
		\hat Q(\boldsymbol{o}, \boldsymbol{a}) = Q(\boldsymbol{o},\boldsymbol{a})+\emph{n}, \\
		\emph{n} \sim {\cal N}(0, \sigma^2).
	\end{equation}
	Then, let $\hat Q_\alpha$ and $\hat Q_\beta$ represent the corresponding shared encrypted local Q-values for the type-$\alpha$ and type-$\beta$ vehicular agents, respectively. 
	Moreover, considering the fact that the type-$\beta$ vehicular agents cannot learn their policies individually due to the unavailability of the rewards, we aim for federatively training the policies for both types of vehicular agents through only sharing the encrypted local Q-values. Thus, the objective of vehicular agents is to find an optimal joint policy for maximizing the expected long-term global reward under local observations and privacy-preserving requirements, which can be formulated as 
		\begin{equation}
			\max_{\pi_\alpha, \pi_\beta}~ \!\!
			\sum_{t=1}^{T}{\mathbb{E}}[\gamma^{t-1} r_{t}|
			\pi_\alpha(\boldsymbol{a}_t^\alpha|\boldsymbol{o}_t^\alpha,r_t,{\hat Q_\beta}),
			\pi_\beta(\boldsymbol{a}_t^\beta|\boldsymbol{o}_t^\beta,{\hat Q_\alpha})],
		\end{equation}
	where $\pi_\alpha$ and $\pi_\beta$ represent the  policies of both types of vehicular agents, respectively.
\begin{algorithm}[H]
	\caption{Federated Multi-Agent Joint Edge Association and Power Allocation Solution}
	\label{alg1}
	\begin{algorithmic}[1]
		\STATE Random initialize $\theta$, $\theta^*$ and $\cal M$. 
		\FOR{episode=$1:E_{max}$}
		\FOR{TS $t=1:T$}
		\FOR{Type-$\alpha$ vehicular agent}
		\STATE Observe $\boldsymbol{o}_t^\alpha$; 
		\FOR{Type-$\beta$ vehicular agent}
		\STATE Observe $\boldsymbol{o}_t^\beta$;
		\STATE Select $\tilde{\boldsymbol{a}}_t^\beta$ with probability $\epsilon$,\\ 
		otherwise $\tilde{\boldsymbol{a}}_t^\beta = \mathop {\arg \max }\limits_{\boldsymbol{a}}{Q_\beta}(\boldsymbol{o}_t^\beta ,\boldsymbol{a} ;\theta_\beta) $;
		\STATE Obtain ${{\hat Q}_\beta }
		\! = \! Q_\beta({\boldsymbol{o}_t^\beta ,\tilde{\boldsymbol{a}}_t^\beta ;{\theta _\beta }}) + \emph{n}, \emph{n} \sim {\cal N}(0, \sigma^2)$.
		\ENDFOR
		\STATE Select joint action $\boldsymbol{a}_t$ with probability $\epsilon$, \\ otherwise ${\boldsymbol{a}_t} = \mathop {\arg \max }\limits_{\boldsymbol{a}} {Q_{\text{MLP}}^\alpha}(\boldsymbol{o}_t^\alpha ,\boldsymbol{a},{{\hat Q}_\beta };{\theta _{\text{MLP}}})$.
		\ENDFOR
		\STATE Decompose the joint action  $\boldsymbol{a}_t$ to  $\boldsymbol{a}_t^\alpha$ and $\boldsymbol{a}_t^\beta$;
		\STATE Execute $\boldsymbol{a}_t^\alpha$ , $\boldsymbol{a}_t^\beta$ and receive $r_t$, $\boldsymbol{o}_{t+1}^\alpha$ and $\boldsymbol{o}_{t+1}^\beta$; 
		\STATE Store $(\boldsymbol{o}_t^\alpha, {\boldsymbol{a}_t^\alpha},r_t, \boldsymbol{o}_{t + 1}^\alpha)$ and $(\boldsymbol{o}_t^\beta, \boldsymbol{a}_t^\beta, {\boldsymbol{o}_{t+1}^\beta})$ into $\cal M$; 
		\STATE Sample mini-batch $\{(\boldsymbol{o}_j^\alpha ,\boldsymbol{a}_j^\alpha,{r_j},\boldsymbol{o}_{j+1}^\alpha)\}_{j=1}^N$ and $\{(\boldsymbol{o}_j^\beta,\boldsymbol{a}_j^\beta)\}_{j=1}^N$ from $\cal M$;
		\STATE Set ${{\hat Q}_\beta }
		= Q_\beta({\boldsymbol{o}_j^\beta, \boldsymbol{a}_j^\beta;{\theta _\beta }} ) + \emph{n}, \emph{n} \sim{\cal N}(0, \sigma^2)$;
		\STATE Update $\theta_\alpha$ and $\theta_{\text{MLP}}$ according to Eqs.~(\ref{10}) and (\ref{11});
		\STATE Set ${{\hat Q}_\alpha } = Q_\alpha({\boldsymbol{o}_j^\alpha ,\boldsymbol{a}_j^\alpha;{\theta _\alpha }}) + \emph{n}, \emph{n} \sim {\cal N}(0, \sigma^2)$; 
		\STATE Update $\theta_\beta$ and $\theta_{\text{MLP}}$ according to Eqs. (\ref{10}) and (\ref{12}).
		\ENDFOR
		\ENDFOR 
	\end{algorithmic} 
\end{algorithm}

    \subsection{Proposed Federated Multi-agent JEAPA Solution}
    As one of the most representative DRL algorithms, a Deep Q Network (DQN) employs DNN-based Q-learning for performing complex function approximation~\cite{15}, hence it has the ability to accurately approximate the value function, when dealing with the high-dimensional observation space. However, from the perspective of privacy preservation, the vehicles can only make decisions based on their own local observations, thus a single-agent DQN that trains a joint policy relying on the global state becomes infeasible.  
    
    To address this issue, we adopt a centralized training and distributed execution (CTDE) framework, where all vehicular agents are trained centrally at the MBS through sharing the local Q-values, and make decisions distributively based on the trained policies through their own local observations. For a type-$\alpha$ vehicular agent, its policy can be obtained directly by interacting with the environment via DQNs, since the global reward knowledge can be obtained. By contrast, owing to the unavailability of the global reward for a type-$\beta$ vehicular agent, its policy cannot be learned from itself. Nevertheless, the encrypted information can be shared among agents. Hence, we can utilize the type-$\beta$ vehicular agent's encrypted local Q-values to assist $\alpha$ for constructing a joint policy. To be specific, as shown in Fig.~1, each agent initially acquires Q-values from the local Q-networks and encrypts them using the Gaussian differential method of \cite{16}. Afterwards, the encrypted local Q-values are shared through a federated Q network, and the joint actions are generated. The details of the framework are as follows:
	\begin{enumerate}[fullwidth,itemindent=1em,label=\arabic*)]
		\item{\emph{Local Q-network}: For type-$\alpha$ and type-$\beta$ vehicular agents, local Q-networks are conceived for estimating the state-action-value function, which are denoted as $Q_{\alpha}(\cdot;{\theta _{\alpha}})$ and $Q_{\beta}(\cdot;{\theta _{\beta}})$, respectively. Herein, $\theta_{\alpha}$ and $\theta_{\beta}$ are the corresponding DNN weights.
		}
		\item{
			\emph{Gaussian differential privacy}: To encrypt the local Q-values for privacy preservation, we adopt the differential privacy method of \cite{16}, where the local Q-values are added with a random Gaussian variable according to Eq.~(\ref{5}).
		}
		\item{
			\emph{Federated Q network}: Given that the input of the federated Q-network is the vector of batch-size concatenated from tabular data, a multilayer perceptron (MLP) \cite{5499042} network can be established to share the encrypted local Q-values and to calculate a global output, denoted as $Q_{\text{MLP}}(\cdot;{\theta_{\text{MLP}}})$, for predicting the joint action, where ${\theta_{\text{MLP}}}$ represents the MLP network weights.
		} 
        \item{
    	\emph{Experience replay}: To improve the stability of RL, an experience replay buffer, denoted as ${\cal M}$, is employed for mitigating the strong correlation between samples. During training, both vehicular agents sample a minibatch $\{(\boldsymbol{o}_j^\alpha ,\boldsymbol{a}_j^\alpha,{r_j},\boldsymbol{o}_{j+1}^\alpha)\}_{j=1}^N$ and $\{(\boldsymbol{o}_j^\beta,\boldsymbol{a}_j^\beta)\}_{j=1}^N$ of $N$ transitions from  ${\cal M}$, respectively, where $r_j$ is the global reward.
        }
        \item{
        \emph{Separate target networks}: For preventing frequent updates and reducing both the divergence as well as oscillation of training, target networks are cloned by the main networks of the local Q-network and MLP network, which are denoted by $Q_{\alpha}^*(\cdot;{\theta_{\alpha}^*})$ and $Q_{\text{MLP}}^*(\cdot;{\theta_{\text{MLP}}^*})$, respectively. Note that the target value of MLP can only be computed by the type-$\alpha$ vehicular agents but then may be shared with the type-$\beta$ vehicular agents, given by
        	\begin{equation}
        		\label{10}
        		{Y_j} = (r_j + \gamma [\mathop {\max }_{{\boldsymbol{a}_{j+1}^{\alpha}}} Q_{_{\text{MLP}}}^\alpha (\boldsymbol{o}_{j + 1}^{\alpha},{\boldsymbol{a}_{j + 1}^\alpha},{{\hat Q}_\beta};{\theta _\alpha^* },\theta _{\text{MLP}}^*)]).
        	\end{equation}
        Moreover, different from the commonly-used FMARL-based solution, which directly updates the weights of the global network by fitting the aggregated DNN weights of local networks, the local Q-networks and the MLP network in our solution are updated by minimizing the loss function through the popular gradient descent method,  represented as 
        	\begin{equation} 
        		\label{11}
        		\begin{aligned}
        			L_j^\alpha({{\theta _\alpha },{\theta _{\text{MLP}}}}) \!=\! \mathbb{E}
        			[({Y_j} \!-\! Q_{_{\text{MLP}}}^\alpha
        			({\boldsymbol{o}_j^\alpha,\boldsymbol{a}_j^\alpha,{\hat Q}_\beta;{\theta _\alpha },{\theta _{\text{MLP}}}}))^2]
        		\end{aligned}
        	\end{equation}and
        	\begin{equation} 
        		\label{12}
        		\begin{aligned}
        			L_j^\beta({{\theta _\beta },{\theta _{\text{MLP}}}}) \!=\! \mathbb{E}
        			[({Y_j} \!-\! Q_{_{\text{MLP}}}^\beta
        			({\boldsymbol{o}_j^\beta,\boldsymbol{a}_j^\beta,{\hat Q}_\alpha;{\theta _\beta },{\theta _{\text{MLP}}}}))^2].
        		\end{aligned}
        	\end{equation}
		}
	\end{enumerate}

    In a nutshell, the training process of the overall workflow is shown in Algorithm \ref{alg1}. Specifically, (i) first type-$\alpha$ vehicular agent initially computes the target value $Y_j$ for updating its own local Q-network and MLP network. Then it computes the encrypted local Q-values ${\hat Q}_\alpha$; (ii) with $Y_j$, $\theta_{\text{MLP}}$ and ${\hat Q}_\alpha$ sent by $\alpha$, the type-$\beta$ vehicular agent updates the networks, and then computes the encrypted local Q-values ${\hat Q}_\beta$ to assist $\alpha$'s model training. As such, when testing, only ${\hat Q}_\alpha$ and ${\hat Q}_\beta$ have to be shared for constructing the joint policy.
	\section{Simulation Results And Evaluations}
	\subsection{Simulation Settings}
	In our simulations, we consider a pair of vehicular agents\footnote{The settings can be extended to more agents by grouping such agent pairs.} driving along the road and 12 RSUs located uniformly along both sides of the road, with the maximum coverage range of $200$~m. The length of road is set as $1$ km and $O_{\max}=4$. Additionally, we adopt the following channel model: the path loss (dB) is $G_t^{k,r}=128.1 + 37.6{\log _{10}}d_t^{k,r}$, where $d_t^{k,r}$ is the distance in \textit{km} between vehicle $k$ and RSU~$r$ at TS $t$; the small-scale fading is Rayleigh fading with unit variance. The transmission power of RSUs is set to [23, 35] dBm and the minimum data rate constraint is set to $8$ bit/s/Hz. The mobility pattern of vehicles follows a Gauss-Markov stochastic process~\cite{17}, where the corresponding asymptotic mean and the standard deviation of each vehicular velocity are set to $[5\text{m/s}, 10\text{m/s}]$ and $0.1$, respectively. Moreover, the memory-depth that characterizes the temporal correlation of vehicular speed is set to $0.1$. The weight factors ${\omega _1}$, ${\omega _2}$ and ${\omega _3}$ are $0.5, 0.25, 0.25$, respectively. The penalty is set as ~-1.
	
	We construct the local Q-network as a three-layer fully connected neural network with $80$ neurons. With regard to the learning configurations, the learning rate attenuates from $0.01$ to $0.001$ and the discount factor $\gamma$ is set to $0.9$. The size of the mini-batch is set up as 32. Moreover, we exploit the $\epsilon$-greedy exploration using $\epsilon=0.1$ and set the standard deviation $\sigma$ in the Gaussian differential privacy to be 1. 
	
	\vspace{-0.2cm}
	\subsection{Performance Evaluation}  
	To evaluate the efficiency of our proposed algorithm, we compare them to the commonly-used baselines\footnote{We assume that Type-$\alpha$ vehicular agents can share the global reward with Type-$\beta$ vehicular agents, so that Type-$\beta$ vehicular agents can learn the policy individually.} as follows:
	\begin{itemize}
		\item{\emph{Centralized DRL (CDRL)} \cite{15}: 
			With the aid of the Double DQN (DDQN) algorithm, all the vehicles are jointly considered as an agent that processes the global state information as its input and yields the joint policy for training and decision making centrally.}
		\item{\emph{Independent MARL (IMARL)} \cite{10}:
			With the aid of the DDQN algorithm, each vehicle acts as an agent to train its own policy and make decisions distributively relying on their own local observations.}
		\item{\emph{Conventional FMARL} \cite{12}: 
			Based on the IMARL, the vehicular agents could upload the weights of the local Q-networks to the cloud center for federated averaging and then download the aggregated weights from the global network to train their policies distributively.}
	\end{itemize}
\begin{figure}[!t]
	\centering
	\includegraphics[width=0.4\textwidth]{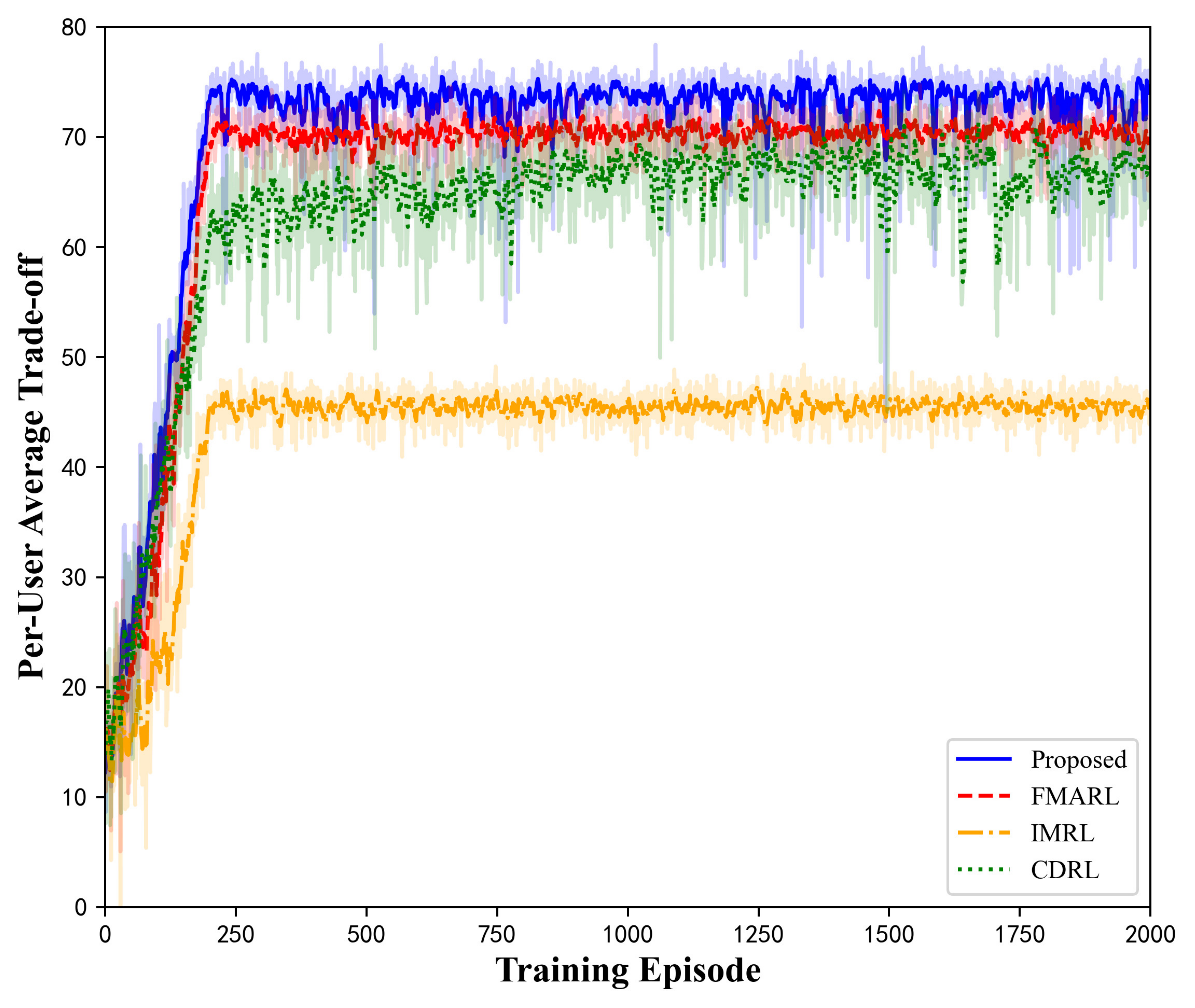}
	\caption{The convergence comparison.}
	\label{fig_2}
\end{figure}
\begin{figure}[!t]
	\centering
	\includegraphics[width=0.4\textwidth]{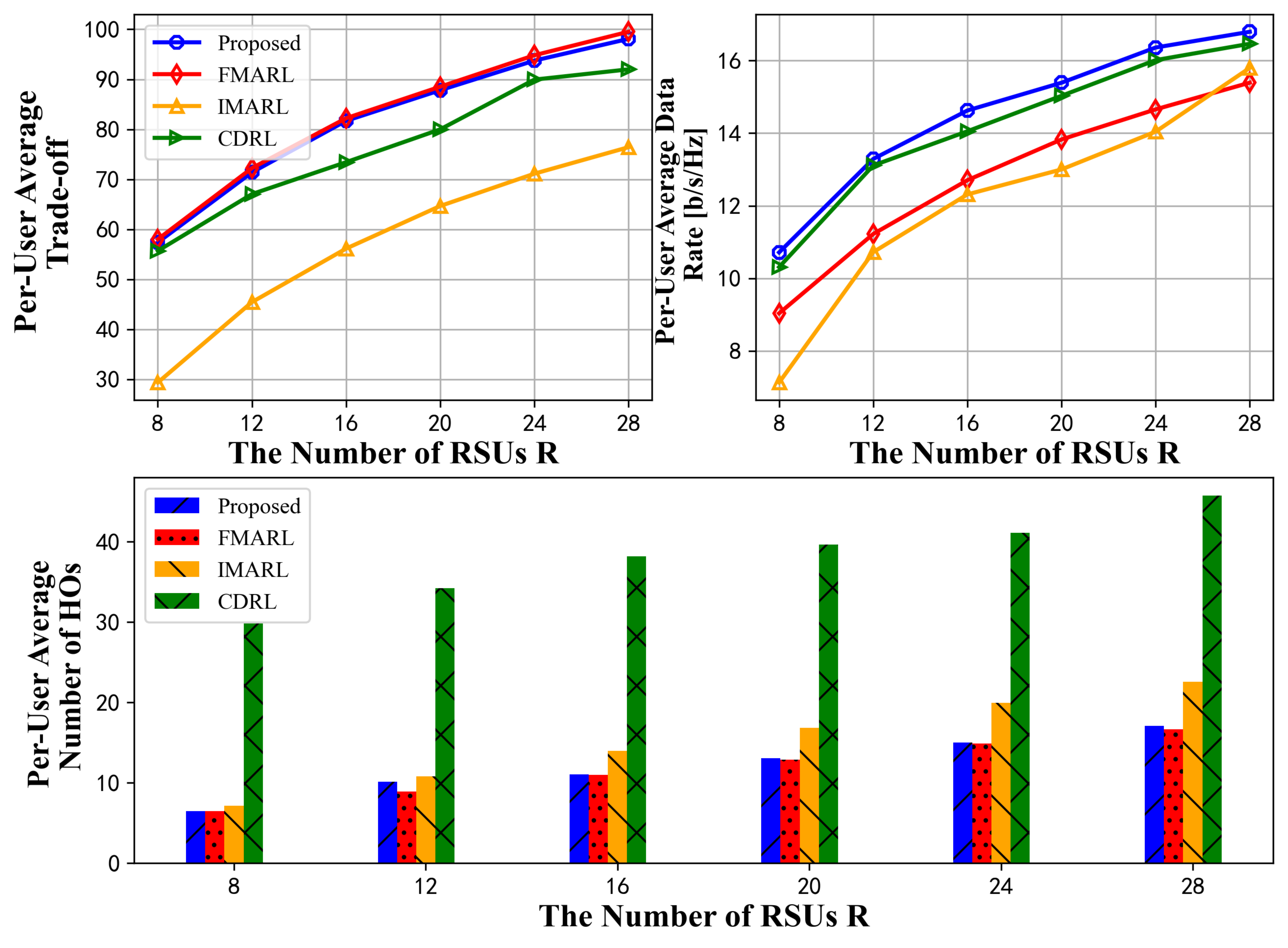}
	\caption{The comparison of per-user average performance versus the number of RSUs $R$.}
	\label{fig_3}
	\vspace{-0.5cm}
\end{figure}

The convergence of all the schemes is illustrated in Fig.~\ref{fig_2}. First, we can observe that the PAT of our proposed algorithm is improving as the training continues and gradually saturates around 250 episodes, which verifies the effectiveness of the proposed algorithm. Next, we can see from Fig.~\ref{fig_2} that the PAT of the proposed algorithm is better than that of the other baselines after convergence, apart from some fluctuations. This implies that sharing the encrypted local Q-values contributes to improving the performance of the learning policy federatively, even though some vehicular agents cannot learn their policies individually.

Fig.~\ref{fig_3} compares the PAT over 100 episodes after convergence versus the number of RSUs. First of all, we can observe from the lower subfigure that the PAT is improving for all solutions upon increasing the number of RSUs. We can also observe a similar trend for the transmission rate in the left subfigure of Fig.~\ref{fig_3}. This is because as the number of RSUs increases, the vehicular agents may have more opportunities for connecting to a closer RSU, thus increasing the data rate. Secondly, the PAT of our proposed scheme is substantially better than that of the CDRL and the IMARL, which is an explicit benefit of the auxiliary training data. Although the FMARL may be slightly better in terms of its PAT than the proposed scheme, the latter achieves a higher privacy-preservation level at the cost of a modest average performance erosion. Moreover, as shown in the right subfigure of Fig.~\ref{fig_3}, our algorithm has a clear performance advantage in optimizing the average data rate. In terms of reducing the average number of HOs in the lower subfigure of Fig.~\ref{fig_3}, our proposed algorithm outperforms the CDRL and the IMARL, and it is slightly inferior to the FMARL, but it has a higher privacy-preservation level. These trends provide evidence again about the explicit benefits of the auxiliary encrypted model training data for learning their policies federatively.

\begin{figure}[!t]
	\centering
	\includegraphics[width=0.4\textwidth]{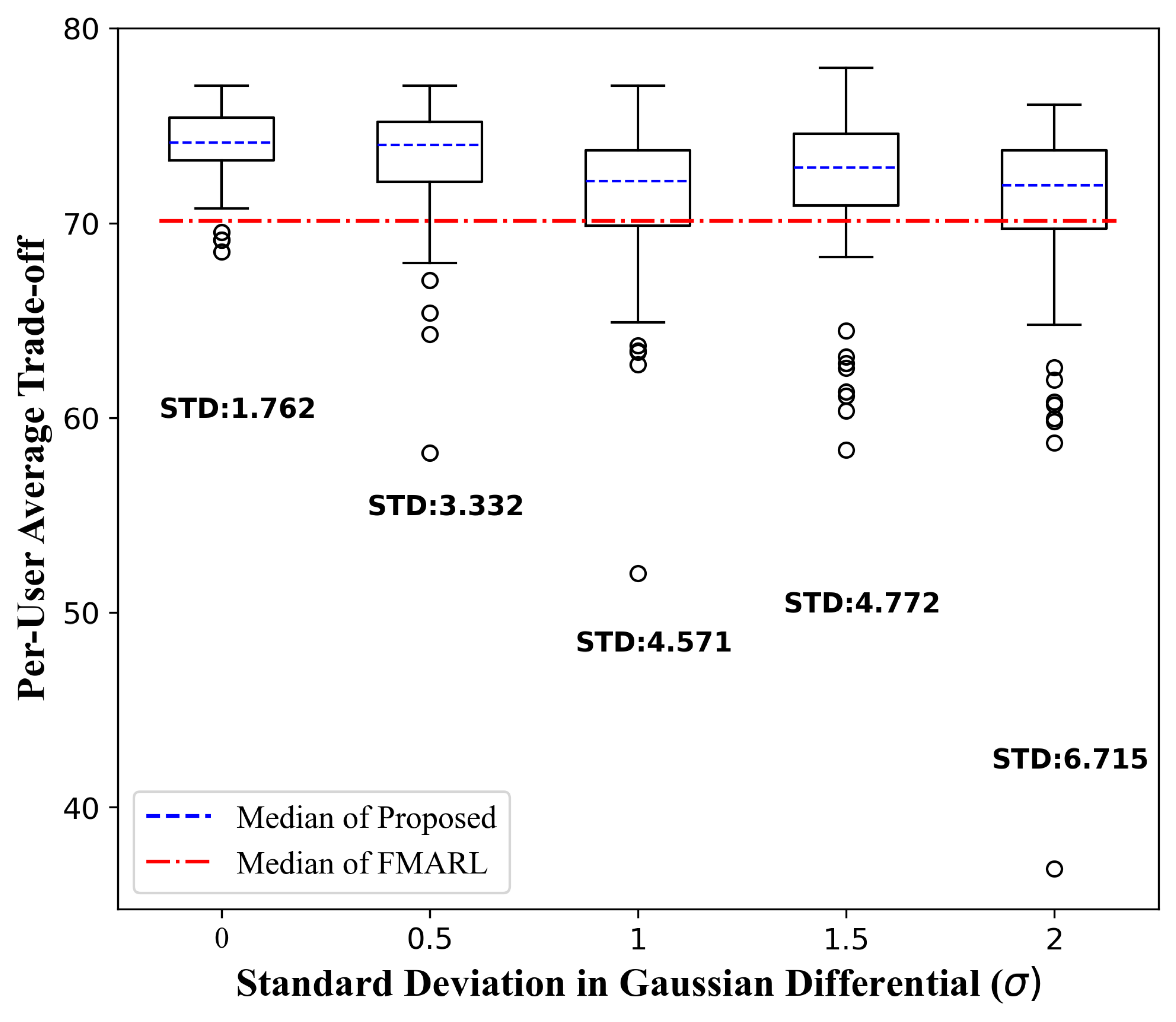}
	\caption{Accuracy versus privacy.}
	\label{fig_4}
\end{figure}

In Fig.~\ref{fig_4}, we investigate the trade-off between the accuracy and the privacy characterized by the standard deviation (SD) $\sigma$ of Gaussian noise added to the shared local Q-values. As shown in Fig.~\ref{fig_4}, with the increase of $\sigma$, the median of the PAT performance tends to decrease. More concretely, the median in the case of $\sigma \neq 0$ is lower than that when $\sigma = 0$. Meanwhile, the SD of the PAT performance is increased as $\sigma$ increases. This is owing to the fact that the Gaussian noise characterizes the lower bound on the expected generalization error that our proposed algorithm can achieve for its decision making. Overall, it can be concluded that a higher privacy-preserving level will lead to lower convergence rate for our proposed algorithm. Furthermore, we can observe that our proposed scheme outperforms the FMARL in terms of the median of the PAT, even though the training data is encrypted for maintaining a higher privacy-preserving level. These results cast a new light on how we strike a compelling trade-off between accuracy and privacy: the FMARL requires all vehicular agents to learn individually and achieves a higher average PAT associated with a lower privacy-preserving level. By contrast, in our proposed scheme some vehicular agents cannot learn individually, but this scheme maintains a higher privacy-preserving level and a higher median PAT.

	\section{Conclusions}
	A federated multi-agent JEAPA framework was conceived for scenarios, when privacy-preserving training is required. By sharing encrypted training data, the privacy of interactions among vehicular agents can be preserved during federative decision-making training. Even if some vehicular agents cannot learn individually, the proposed solution improved our performance metrics and striked a compelling accuracy-privacy trade-off. Our future work will consider 1) the impact of the vehicles' density; 2) the dual function of communicating and computing for RSUs; 3) the application of policy-based cooperative multi-agent RL methods.

\bibliographystyle{IEEEtran}
\def\bibfont{\fontsize{7}{8}\selectfont}
\bibliography{ref}
\end{document}